\documentclass[runningheads,a4paper]{llncs}

\usepackage{amssymb}
\usepackage{graphicx}

\usepackage{url}

\urldef{\mailsc}\path|noswa023@uottawa.ca|    
\newcommand{\keywords}[1]{\par\addvspace\baselineskip
\noindent\keywordname\enspace\ignorespaces#1}

\begin{document}

\mainmatter  

\title{Predicting Rainfall using Machine Learning Techniques
}


%
%
\author{Nikhil Oswal}
%


\institute{School of Electrical Engineering and Computer Science (EECS),\\ University of Ottawa,\\
Ottawa, Canada\\
\mailsc
}

\maketitle

\begin{abstract}
Rainfall prediction is one of the challenging and uncertain tasks which has a significant impact on human society. Timely and accurate predictions can help to proactively reduce human and financial loss. This study presents a set of experiments which involve the use of
prevalent machine learning techniques to build
models to predict whether it is going to rain tomorrow or not based on weather data for that particular day in major cities of  Australia. This comparative study is conducted concentrating on three
aspects: modeling inputs, modeling methods, and pre-processing techniques. The results provide a comparison of various evaluation metrics of these machine learning techniques and their
reliability to predict the rainfall by analyzing the weather data.

\end{abstract}

\section{Introduction}

\noindent Rainfall prediction remains a serious concern and has attracted the attention
of governments, industries, risk management entities, as well as the scientific
community. Rainfall is a climatic factor that affects many human activities like
agricultural production, construction, power generation, forestry and tourism,
among others \cite{1}. To this extent, rainfall prediction is essential since this variable is the one with the highest correlation with adverse natural events such
as landslides, flooding, mass movements and avalanches. These incidents have affected society for years \cite{2}. Therefore, having an appropriate approach for rainfall prediction makes it possible to take preventive and mitigation measures for
these natural phenomena \cite{3}. 

To solve this uncertainty, we used various machine learning techniques and models to make accurate and timely predictions. These paper aims to provide end to end machine learning life cycle right from Data preprocessing to implementing models to evaluating them. Data Preprocessing steps include imputing missing values, feature transformation, encoding categorical features, feature scaling and feature selection. We implemented models such as Logistic Regression, Decision Tree, K Nearest Neighbour, Rule-based and Ensembles. For evaluation purpose, we used Accuracy, Precision, Recall, F-Score and Area Under Curve as evaluation metrics. For our experiments, we train our classifiers using
Australian weather data gathered from various weather stations in Australia.
    
The paper is organized as follows. First, we describe the data set under consideration in Section
2. The adopted methods and techniques are presented in Section 3, while the experiments and results are shown and
discussed in Section 4. Finally, closing conclusions are drawn (Section 5).

\section{Case Study} 
In this paper, the data set under consideration contains daily weather observations from numerous Australian weather stations. The target variable is RainTomorrow which means: Did it rain the next day? Yes or No. The dataset is available at \url{https://www.kaggle.com/jsphyg/weather-dataset-rattle-package} and definitions are adapted from \url{http://www.bom.gov.au/climate/dwo/IDCJDW0000.shtml}.

The data set consists of 23 features and 142k instances. Below are the features.

\begin{table}[!hbt]
\caption{Data set Description} 
\centering 
\begin{tabular}{c c c c} 
\hline\hline 
Feature & Description  \\ [0.5ex] 
\hline 

Date & The date of observation \\
Location & The common name of the location of the weather station \\
MinTemp & The minimum temperature in degrees celsius \\
MaxTemp & The maximum temperature in degrees celsius \\
Rainfall & The amount of rainfall recorded for the day in mm \\
Evaporation & The so-called Class A pan evaporation (mm) in the 24 hours to 9am \\
Sunshine & The number of hours of bright sunshine in the day. \\
WindGustDir & The direction of the strongest wind gust in the 24 hours to midnight \\
WindGustSpeed & The speed (km/h) of the strongest wind gust in the 24 hours to midnight \\
WindDir9am & Direction of the wind at 9am \\
WindDir3pm & Direction of the wind at 3pm \\
WindSpeed9am & Wind speed (km/hr) averaged over 10 minutes prior to 9am \\
WindSpeed3pm & Wind speed (km/hr) averaged over 10 minutes prior to 3pm \\
Humidity9am & Humidity (percent) at 9am \\
Humidity3pm & Humidity (percent) at 3pm \\
Pressure9am & Atmospheric pressure (hpa) reduced to mean sea level at 9am \\
Pressure3pm & Atmospheric pressure (hpa) reduced to mean sea level at 3pm \\
Cloud9am & Fraction of sky obscured by cloud at 9am. \\
Cloud3pm & Fraction of sky obscured by cloud at 3pm.  \\
Temp9am & Temperature (degrees C) at 9am \\
Temp3pm & Temperature (degrees C) at 3pm \\
RainToday & 1 if precipitation exceeds 1mm, otherwise 0 \\
RISK\_MM & The amount of next day rain in mm.  \\
RainTomorrow & The target variable. Did it rain tomorrow? \\ [1ex] 
\hline 
\end{tabular}
\label{table:nonlin} 
\end{table}

\section{Methodology}
In this paper, the overall architecture include four major components: Data Exploration and Analysis, Data Pre-processing, Model Implementation, and Model Evaluation, as shown in Fig. 1.

\begin{figure}[hbt!]
\centering
    \frame{\includegraphics[width=\columnwidth]{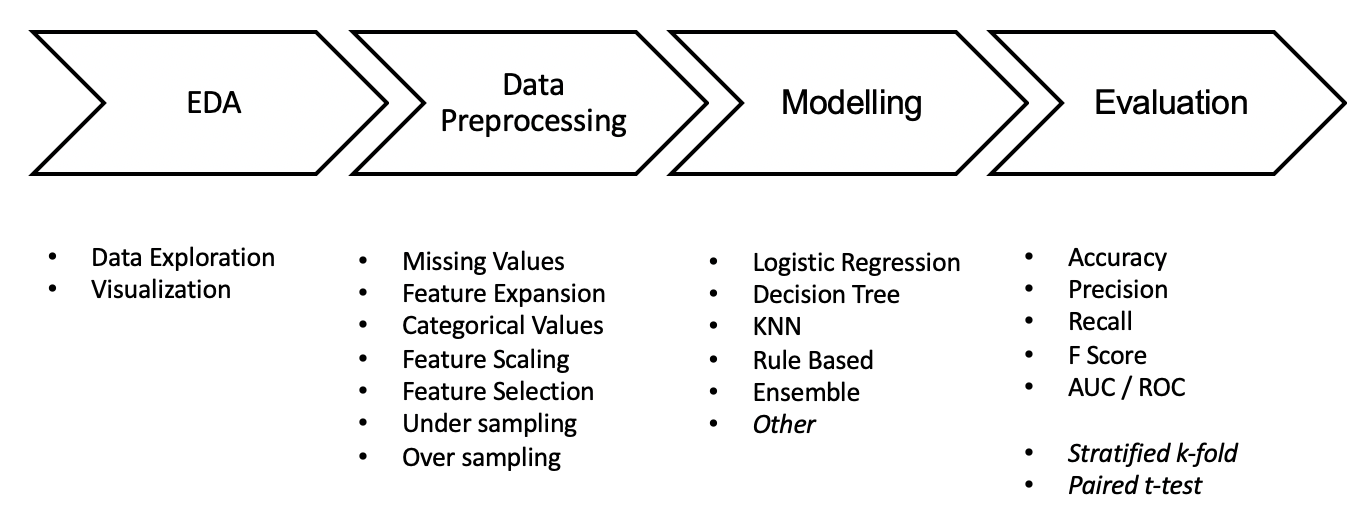}}
    \caption{Overall Architecture.} 
\end{figure}

\subsection{Data Exploration and Analysis}
Exploratory Data Analysis is valuable to machine learning problems since it allows to get closer to the certainty that the future results will be valid, correctly interpreted, and applicable to the desired business contexts \cite{4}. Such level of certainty can be achieved only after raw data is validated and checked for anomalies, ensuring that the data set was collected without errors. EDA also helps to find insights that were not evident or worth investigating to business stakeholders and researchers.

We performed EDA using two methods - 
\textbf{Univariate Visualization} which provides summary statistics for each field in the raw data set (figure 2) and \textbf{Pair-wise Correlation Matrix} which is performed to understand interactions between different fields in the data set (figure 3). 

\begin{figure}[hbt!]
\centering
    \frame{\includegraphics[width=\columnwidth]{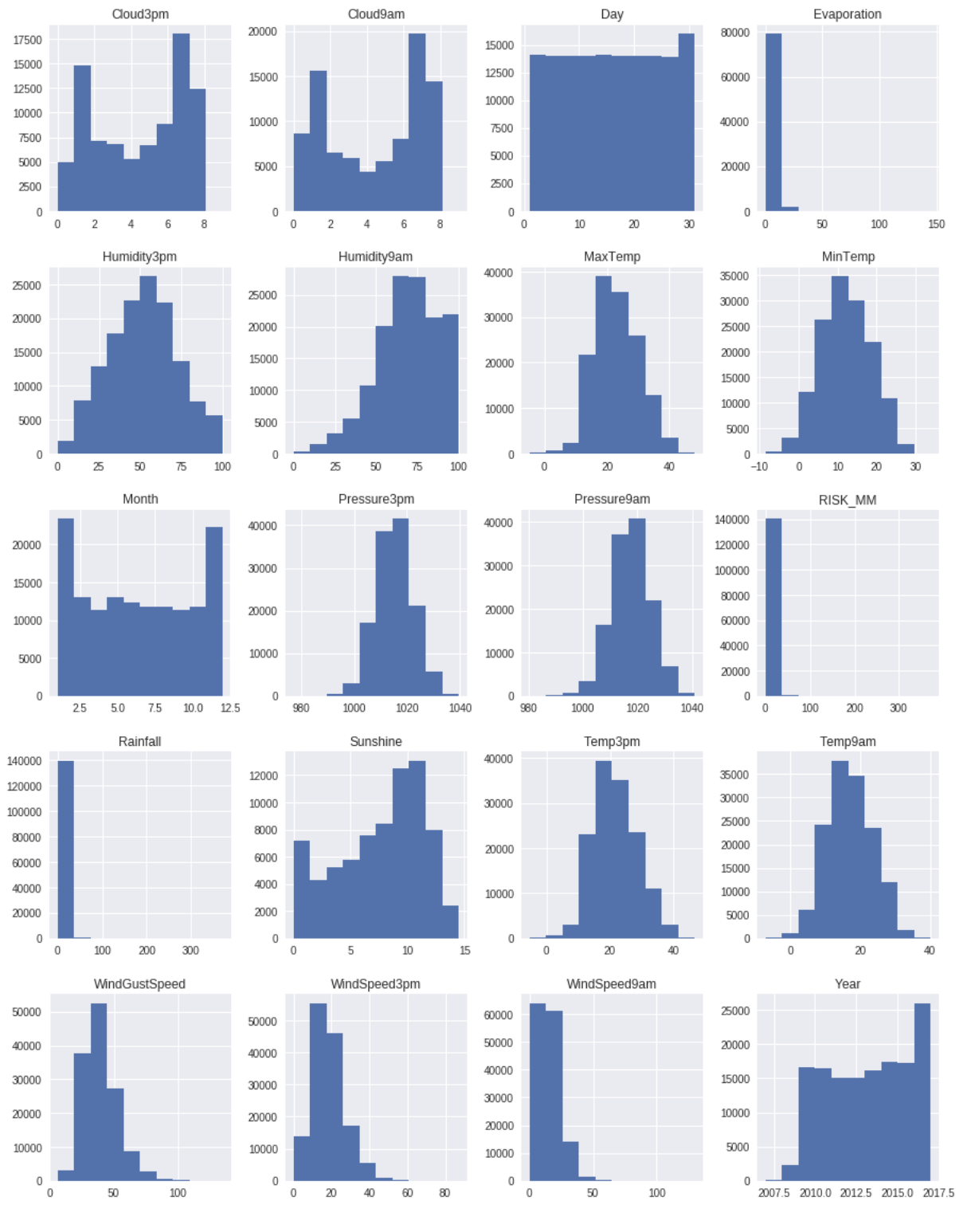}}
    \caption{Univariate Visualization.} 
\end{figure}
\begin{figure}[hbt!]
\centering
    \frame{\includegraphics[width=\columnwidth]{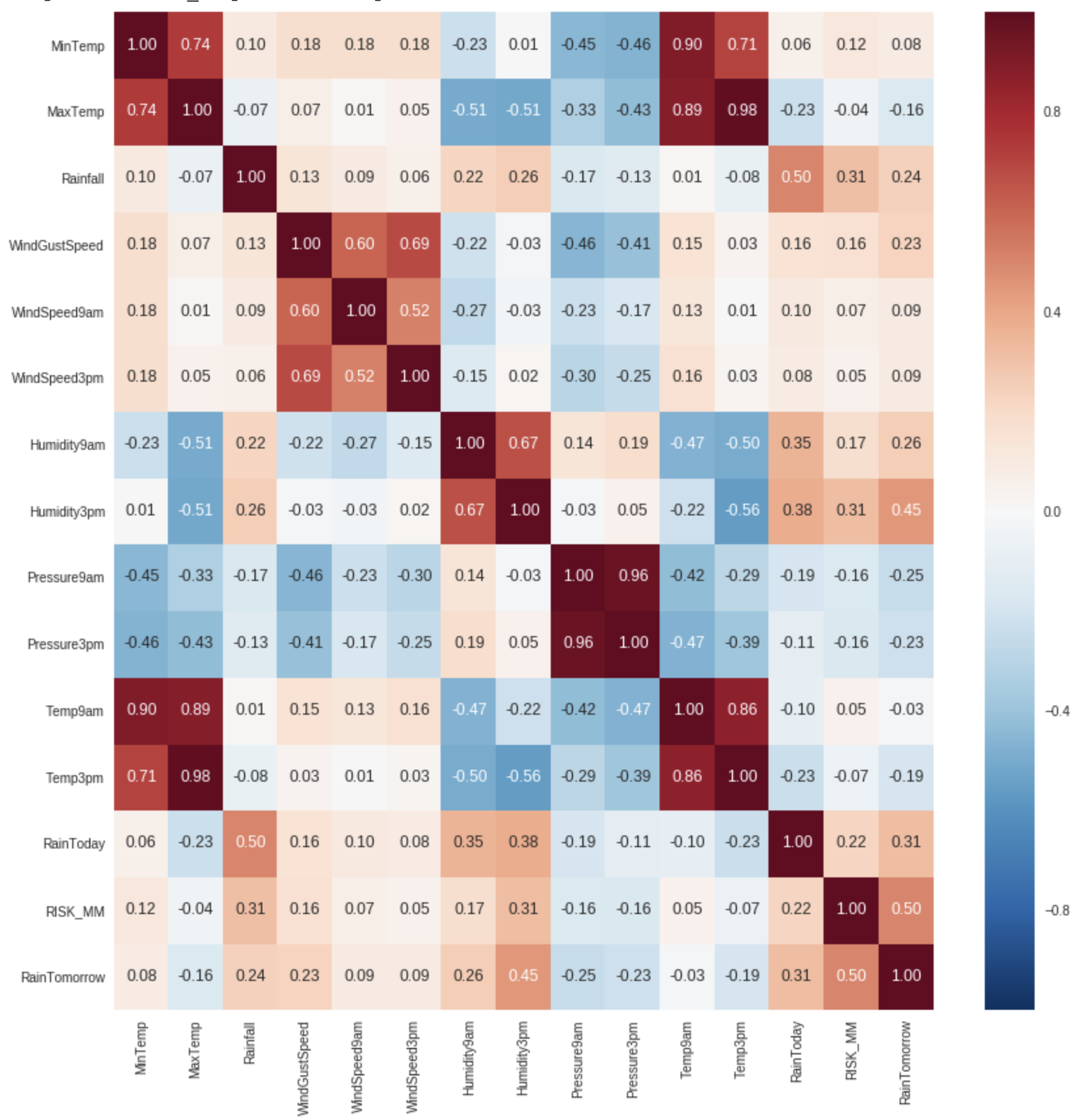}}
    \caption{Heat Map.} 
\end{figure}

\begin{table}[!hbt]
\caption{Irrelevant Features} 
\centering 
\begin{tabular}{c c c c} 
\hline\hline 
Feature & \% of Null values \\ [0.5ex] 
\hline 
Sunshine       &   43\% \\
Evaporation    &   48\% \\
Cloud3pm        &  40\% \\ 
Cloud9am         & 38\% \\[1ex] 
\hline 
\end{tabular}
\label{table:nonlin} 
\end{table}

We have other features with null values too which we will be imputing in our preprocessing steps. If we look the distribution of our target variable, it is clear that we have a class imbalance problem with number of positive instances - 110316 and number of negative instances - 31877. \\
\\
The correlation matrix depicts that the features - MaxTemp, Pressure9am, Pressure3pm, Temp3pm and Temp9am are negatively correlated with target variable. Hence, we can drop this features in our feature selection step later.

\subsection{Data Preprocessing}
Data preprocessing is a data mining technique that involves transforming raw data into an understandable format. Real-world data is often incomplete, inconsistent, and/or lacking in certain behaviors or trends, and is likely to contain many errors. We have carried below preprocessing steps.
\subsubsection{Missing Values:}
As per our EDA step, we learned that we have few instances with null values. Hence, this becomes one of the important step. To impute the missing values, we will group our instances based on the location and date and thereby replace the null values by there respective mean values.
\subsubsection{Feature Expansion:}
Date feature can be expanded to Day, Month and Year and then these newly created features can be further used for other preprocessing steps.
\subsubsection{Categorical Values:}
Categorical feature is one that has two or more categories, but there is no intrinsic ordering to the categories. We have a few categorical features - WindGustDir, WindDir9am, WindDir3pm with 16 unique values. Now it gets complicated for machines to understand texts and process them, rather than numbers, since the models are based on mathematical equations and calculations. Therefore, we have to encode the categorical data. We here tried two different techniques. 
\begin{itemize}
\item \textbf{Dummy Variables}: A Dummy variable is an artificial variable created to represent an attribute with two or more distinct categories/levels \cite{5}. However, as we have 16 unique values, our one feature will now get transformed to 16 new features which in turn results in \textbf{curse of dimensionality}. For each instance, we will have a feature with 1 value and rest 15 features with 0 values.

Example: Categorical Encoding of feature - \textbf{windDir3pm} using Dummy Variables

\begin{figure}[hbt!]
\centering
    \frame{\includegraphics[width=\columnwidth]{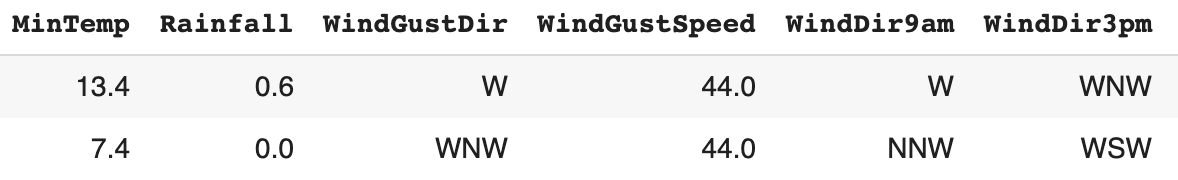}}
    \caption{Sample Instance.} 
\end{figure}
\begin{figure}[hbt!]
\centering
    \frame{\includegraphics[width=\columnwidth]{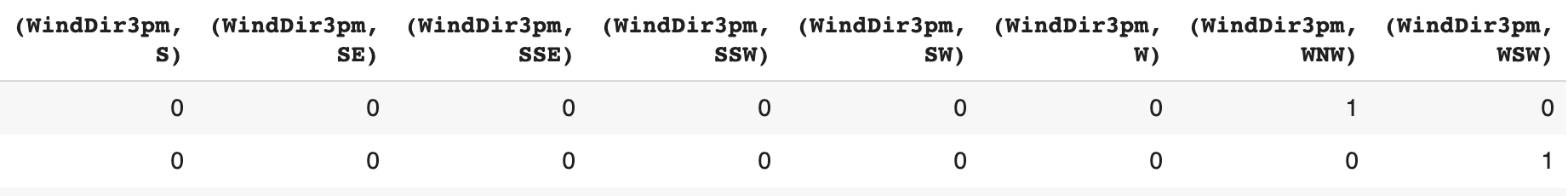}}
    \caption{Dummy Variables.} 
\end{figure}

\item \textbf{Feature Hashing}: Feature hashing scheme is another useful feature engineering scheme for dealing with large scale categorical features. In this scheme, a hash function is typically used with the number of encoded features pre-set (as a vector of pre-defined length) such that the hashed values of the features are used as indices in this pre-defined vector and values are updated accordingly \cite{6}. 

Example: Categorical Encoding of feature - \textbf{windDir3pm} using Feature Hashing
\begin{figure}[hbt!]
\centering
    \frame{\includegraphics[width=\columnwidth]{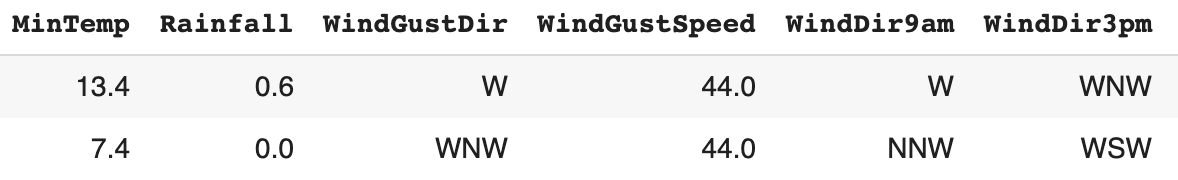}}
    \caption{Sample Instance.} 
\end{figure}
\begin{figure}[hbt!]
\centering
    \frame{\includegraphics[width=\columnwidth]{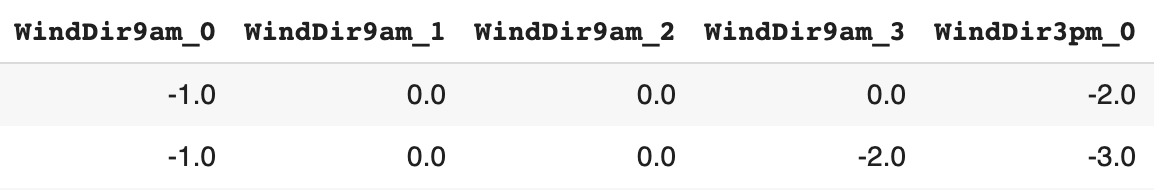}}
    \caption{Feature Hashing.} 
\end{figure}
\end{itemize}
\subsubsection{Feature Scaling:}
Our data set contains features with highly varying magnitudes and range. But since, most of the machine learning algorithms use Euclidean distance between two data points in their computations, this is a problem. The features with high magnitudes will weigh in a lot more in the distance calculations than features with low magnitudes. To suppress this effect, we need to bring all features to the same level of magnitudes. This can be achieved by scaling. We did this using scikit learn's min-max scalar and brought all the features in the range of 0 to 1 \cite{7}.
\begin{figure}[hbt!]
\centering
    \frame{\includegraphics[width=\columnwidth]{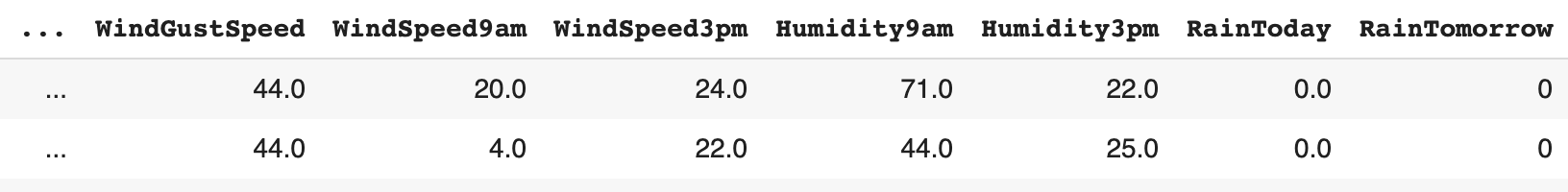}}
    \caption{Sample Instance before scaling.} 
\end{figure}
\begin{figure}[hbt!]
\centering
    \frame{\includegraphics[width=\columnwidth]{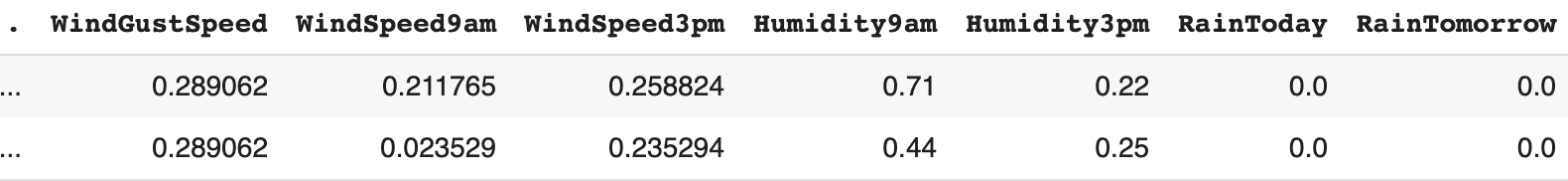}}
    \caption{Sample Instance after scaling.} 
\end{figure}
\subsubsection{Feature Selection}
Feature Selection is the process where you automatically or manually select those features which contribute most to our prediction variable or output. Having irrelevant features in data can decrease the accuracy of the models and make the model learn based on irrelevant features. Feature selection helps to reduce over fitting, improves accuracy and reduces training time. \\
We used two techniques to perform this activity and got the same results.
\begin{itemize}
    \item \textbf{Univariate Selection}: Statistical tests can be used to select those features that have the strongest relationship with the output variable. The scikit-learn library provides the SelectKBest class that can be used with a suite of different statistical tests to select a specific number of features. We used chi-squared statistical test for non-negative features to select 5 of the best features from our data set \cite{8} \cite{9}.
    \item \textbf{Correlation Matrix with Heatmap}:
Correlation states how the features are related to each other or the target variable. Correlation can be positive (increase in one value of feature increases the value of the target variable) or negative (increase in one value of feature decreases the value of the target variable). Heatmap makes it easy to identify which features are most related to the target variable, we plotted heatmap of correlated features using the seaborn library (figure 3) \cite{9}.
\end{itemize}

\subsubsection{Handling Class Imbalance}
We learned in our EDA step that our data set is highly imbalanced. Imbalanced data results in biased results as our model doesn't learn much about the minority class. We performed two experiments one with oversampled data and another with undersampled data.
\begin{itemize}
    \item \textbf{Undersampling}: We used Imblearn's random under sampler library to eliminate instances of majority class \cite{10}. This elimination is based on distance so that there is minimum loss of information (figure 10)
    \begin{figure}
    \centering
    \frame{\includegraphics[width=\columnwidth]{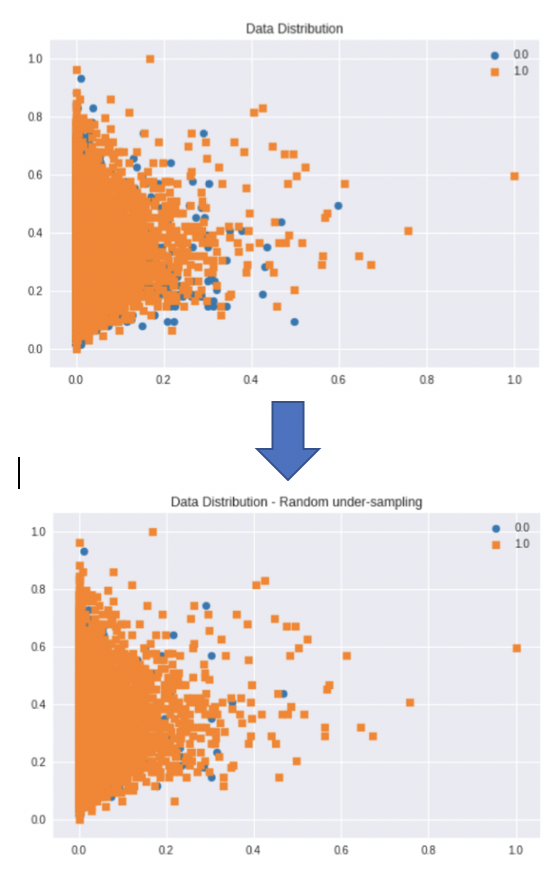}}
    \caption{Undersampling.} 
\end{figure}

    \item \textbf{Oversampling}: We used Imblearn's SMOTE technique to generate synthetic instances for minority class \cite{10}. A subset of data is taken from the minority class as an example and then new synthetic similar instances are created. (figure 11)
    
    \begin{figure}
    \centering
    \frame{\includegraphics[width=\columnwidth]{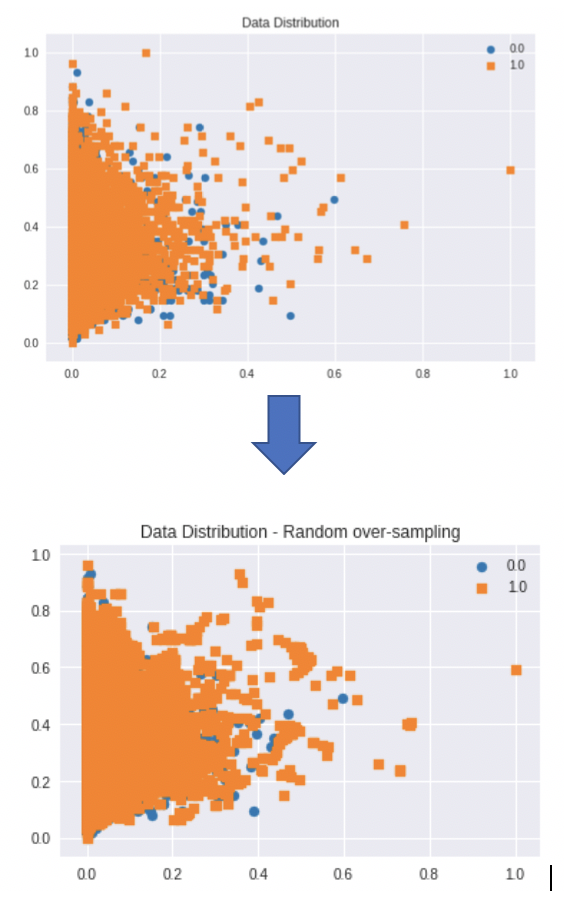}}
    \caption{Oversampling.} 
\end{figure}
    
\end{itemize}
\subsection{Models}

We chose different classifiers each belonging to different model family (such as Linear classifier, Tree-based, Distance-based, Rule-based and Ensemble). All the classifiers were implemented using \textbf{scikit-learn} except for Decision table which was implemented using \textbf{weka}.

The following classification algorithms have been
used to build prediction models to perform the
experiments: 

\subsubsection{Logistic Regression} is a classification algorithm used to predict a binary outcome (1 / 0, Yes / No, True / False) given a set of independent variables. To represent binary / categorical outcome, we use dummy variables. We can also think of logistic regression as a special case of linear regression when the outcome variable is categorical, where we are using log of odds as dependent variable. In simple words, it predicts the probability of occurrence of an event by fitting data to a logit function. Hence, this makes Logistic Regression a better fit as ours is a binary classification problem.
\subsubsection{Decision Tree} have a natural “if … then … else …” construction that makes it fit easily into a programmatic structure. They also are well suited to categorization problems where attributes or features are systematically checked to determine a final category. It works for both categorical and continuous input and output variables. In this technique, we split the population or sample into two or more homogeneous sets (or sub-populations) based on most significant splitter / differentiator in input variables. This characteristics of Decision Tree makes it a good fit for our problem as our target variable is binary categorical variable.
\subsubsection{K - Nearest Neighbour} is a non-parametric and lazy learning algorithm. Non-parametric means there is no assumption for underlying data distribution. In other words, the model structure is determined from the dataset. Lazy algorithm means it does not need any training data points for model generation. All training data used in the testing phase. KNN performs better with a lower number of features than a large number of features. We can say that when the number of features increases than it requires more data. Increase in dimension also leads to the problem of overfitting. However, we have performed feature selection which helps to reduce dimension and hence KNN looks a good candidate for our problem. 

Our Model's configuration: We tried various values of n ranging from 3 to 30 and learned that the model performs best with n as \textbf{25}, \textbf{27} and \textbf{29}.
\subsubsection{Decision table} provides a handy and compact way to represent complex business logic. In a decision table, business logic is well divided into conditions, actions (decisions) and rules for representing the various components that form the business logic. \cite{11} This was implemented using Weka.
\subsubsection{Random Forest} is a supervised ensemble learning algorithm. ‘Ensemble’ means that it takes a bunch of ‘weak learners’ and have them work together to form one strong predictor. Here, we have a collection of decision trees, known as “Forest”. To classify a new object based on attributes, each tree gives a classification and we say the tree “votes” for that class. The forest chooses the classification having the most votes (over all the trees in the forest). 

Our Model's configuration: number of weak learners = 100, maximum depth of each tree = 4
\subsubsection{AdaBoost} fits a sequence of weak learners on different weighted training data. It starts by predicting original data set and gives equal weight to each observation. If prediction is incorrect using the first learner, then it gives higher weight to observations which have been predicted incorrectly. Being an iterative process, it continues to add learner(s) until a limit is reached in the number of models or accuracy. 

Our Model's configuration: number of weak learners = 50

\subsubsection{Gradient Boosting} Here, many models are trained sequentially. Each new model gradually minimizes the loss function (y = ax + b + e, where ‘e’ is the error term) of the whole system using Gradient Descent method. The learning method consecutively fits new models to give a more accurate estimate of the response variable. The main idea behind this algorithm is to construct new base learners which can be optimally correlated with negative gradient of the loss function, relevant to the whole ensemble. 

Our Model's configuration: number of weak learners = 100, learning rate = [0.05, 0.1, 0.25], maximum features = 2, maximum depth = 2

\subsection{Evaluation}
For evaluating our classifiers we used below evaluation metrics \cite{12}.

\subsubsection{Accuracy} is the ratio of number of correct predictions to the total number of input samples. It works well only if there are equal number of samples belonging to each class. As we have, imbalanced data, we will also consider other metrics.

\subsubsection{Area Under Curve(AUC)} is used for binary classification problem. AUC of a classifier is equal to the probability that the classifier will rank a randomly chosen positive example higher than a randomly chosen negative example

\subsubsection{Precision} is the number of correct positive results divided by the number of positive results predicted by the classifier.

\subsubsection{Recall} is the number of correct positive results divided by the number of all relevant samples (all samples that should have been identified as positive).

\subsubsection{F1 Score} is the Harmonic Mean between precision and recall. The range for F1 Score is [0, 1]. It tells how precise our classifier is (how many instances it classifies correctly), as well as how robust it is (it does not miss a significant number of instances). High precision but lower recall, gives you an extremely accurate, but it then misses a large number of instances that are difficult to classify. The greater the F1 Score, the better is the performance of our model.

\subsubsection{Confusion Matrix} gives us a matrix as output and describes the complete performance of the model. It focuses  on \textbf{True Positives} - the cases in which we predicted YES and the actual output was also YES; \textbf{True Negatives} - the cases in which we predicted NO and the actual output was NO; \textbf{False Positives} - the cases in which we predicted YES and the actual output was NO;
 \textbf{False Negatives} - the cases in which we predicted NO and the actual output was YES.

\subsubsection{Stratified k-fold} As our data is imbalanced, we trained our models using a stratified k-fold approach where the data is divided into k folds each of equal proportion of positives and negatives. These metrics would be more reliable and less biased. We used the value of k as 10.

\subsubsection{Statistical testing} For the purpose of comparing the performance of different classifiers, we performed paired t testing among the top three classifiers.

\section{Experiments and Results}

For all the experiments and development of classifiers, we used Python 3 and Google colab's Jupyter Notebook. We used libraries such as Sckit Learn, Matplotlib, Seaborn, Pandas, Numpy and Imblearn. We used weka for implementing Decision Table.

We carried experiments with different input data; one with the original dataset, then with the undersampled dataset and last one with the oversampled dataset. We splitted out dataset in ratio of 75:25 for training and testing purpose.

\subsubsection{Experiment 1 - Original Dataset:}
Post all the preprocessing steps (as mentioned above in the Methodology section), we ran all the implemented classifiers each one with the same input data (Shape: 92037 x 4). Figure 12 depicts two considered metrics (10-skfold Accuracy and Area Under Curve) for all the classifiers.

\begin{figure}
\centering
\frame{\includegraphics[width=\columnwidth]{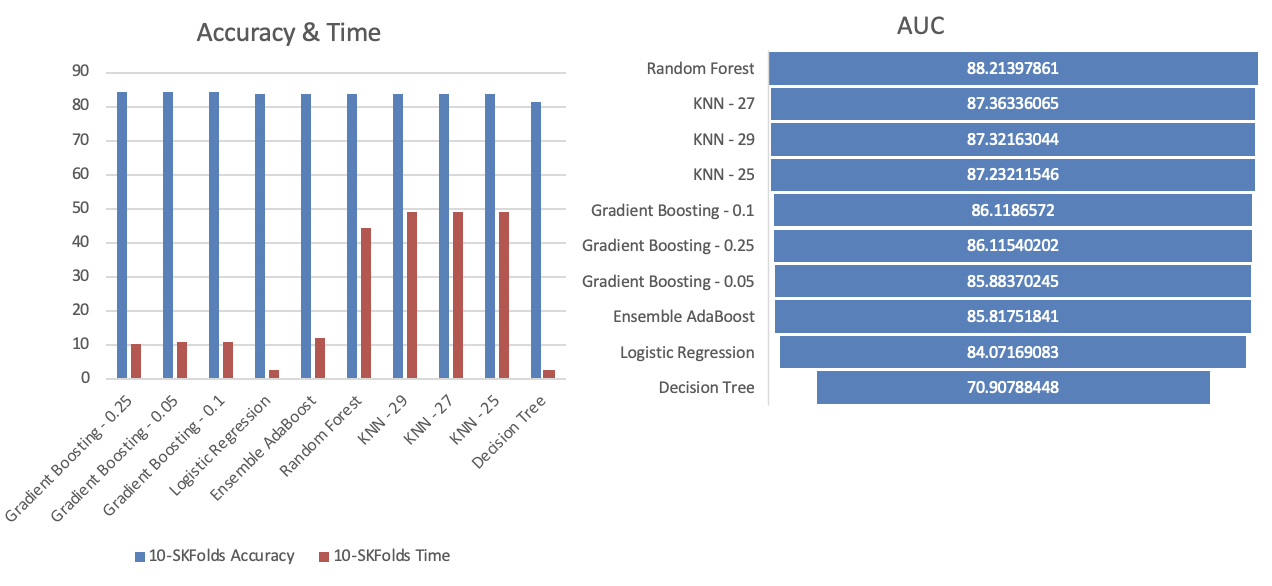}}
\caption{Experiment 1.} 
\end{figure}

Accuracy wise Gradient Boosting with a learning rate of 0.25 performed best, coverage wise Random Forest and Decision Tree performed worsts.

\subsubsection{Experiment 2 - Undersampled Dataset:}
Post all the preprocessing steps (as mentioned above in the Methodology section) including the undersampling step, we ran all the implemented classifiers each one with the same input data (Shape: 54274 x 4). Figure 13 depicts two considered metrics (10-skfold Accuracy and Area Under Curve) for all the classifiers.
\begin{figure}
\centering
\frame{\includegraphics[width=\columnwidth]{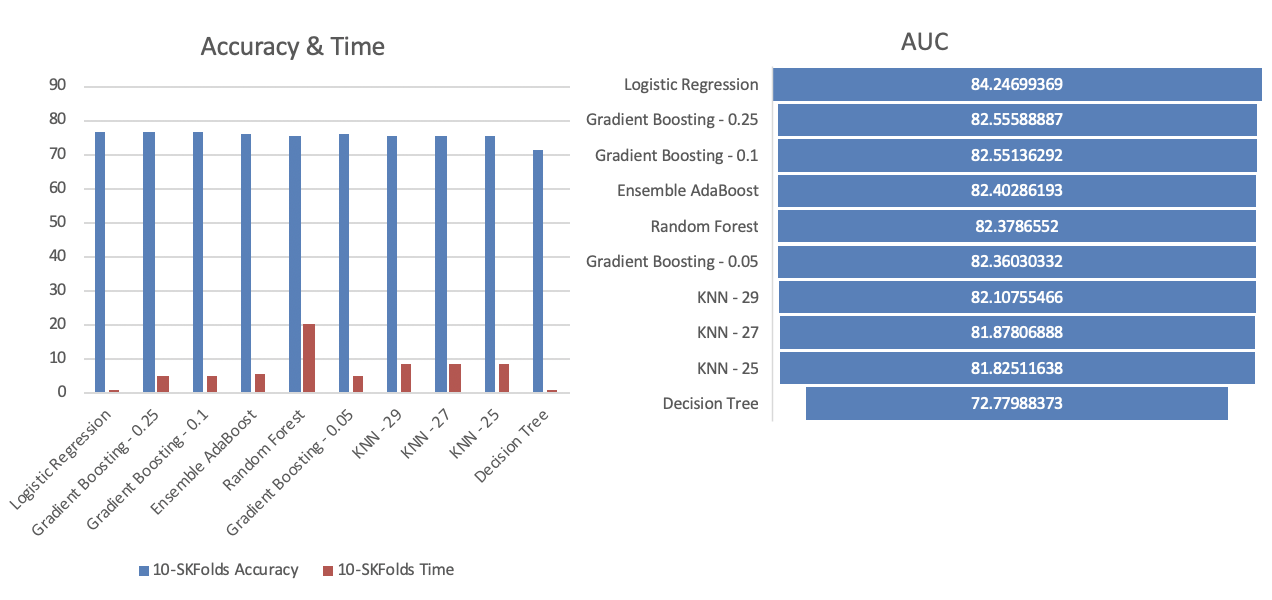}}
\caption{Experiment 2.} 
\end{figure}

Accuracy and coverage wise Logistic Regression performed best and Decision Tree performed worsts.

\subsubsection{Experiment 3 - Oversampled Dataset:}
Post all the preprocessing steps (as mentioned above in the Methodology section) including the oversampling step, we ran all the implemented classifiers each one with the same input data (Shape: 191160 x 4). Figure 14 depicts two considered metrics (10-skfold Accuracy and Area Under Curve) for all the classifiers.

\begin{figure}
\centering
\frame{\includegraphics[width=\columnwidth]{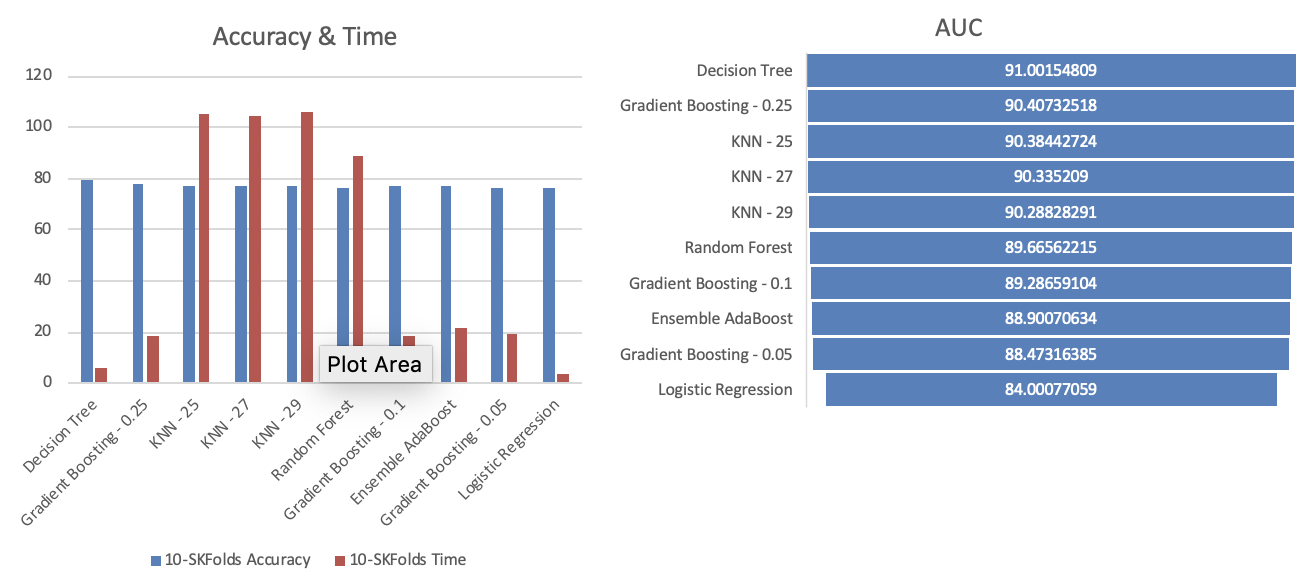}}
\caption{Experiment 3.} 
\end{figure}

Accuracy and coverage wise Decision Tree  performed best and Logistic Regression  performed worsts.

We have varying range of results with respect to different input data and different classifiers. Other metrics are followed in appendix.

\section{Discussion}

With the issues with our original dataset, we learned many things considering all the preprocessing steps that we carried to rectify them. The first important thing we learned is the importance of knowing your data. While imputing the missing value, we grouped two other features and calculated the mean instead of directly calculating the mean for all the instances. This way our imputed values were closer to the correct information. Another thing we learned is about the leaky features. While exploring our data, we came to that one of our feature (RiskMM) was used for generating the target variable and hence it made no sense to use this feature for predictions.

We learned about the curse of dimensionality while dealing with categorical variables which we solved using feature hashing. We also learned two techniques for performing feature selection - univariate selection and correlation heat map. We also explore undersampling and oversampling techniques while handling the class imbalance problem. 

With the experiments that we carried using different data, we also came to know that in a few cases we have achieved higher accuracy (Decision Tree) clearly implying the classic case of overfitting. We also observed that the performance of classifiers varied with different input data. To count a few, Logistic Regression performed best with undersampled data whereas it performed worst with oversampled data; same goes with KNN, it performed best with oversampled data and worst with undersampled data. Hence we can say that the input data has a very important role here. Ensembles to be precise Gradient Boosting performed pretty consistently in all the experiments.

\section{Conclusion and Future Work}
In this paper, we explored and applied several preprocessing steps and learned there impact on the overall performance of our classifiers. We also carried a comparative study of all the classifiers with different input data and observed how the input data can affect the model predictions. 

We can conclude that Australian weather is uncertain and there is no such correlation among rainfall and the respective region and time. We figured certain patterns and relationships among data which helped in determining important features. Refer to the appendix section.

As we have a huge amount of data, we can apply Deep Learning models such as Multilayer Perceptron, Convolutional Neural Network, and others. It would be great to perform a comparative study between the Machine learning classifiers and Deep learning models.

\section*{Appendix}

\noindent Knowledge / Pattern discovered from our dataset.

\begin{figure}
\centering
\frame{\includegraphics{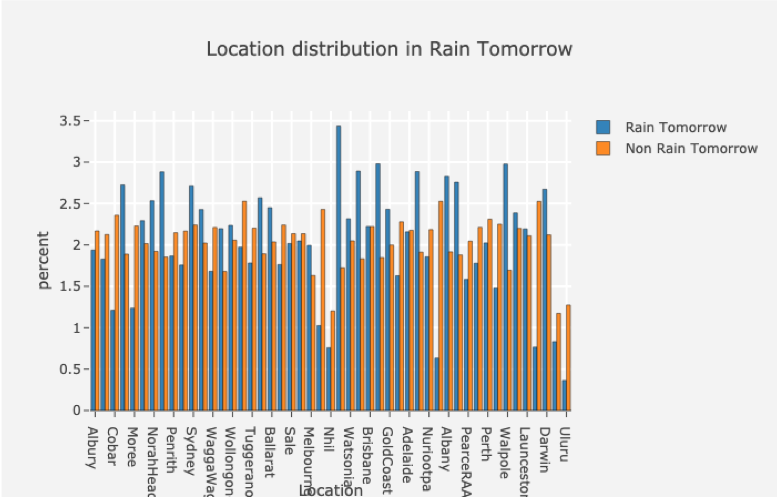}}
\caption{Location and Rainfall.} 
\end{figure}

\begin{figure}
\centering
\frame{\includegraphics{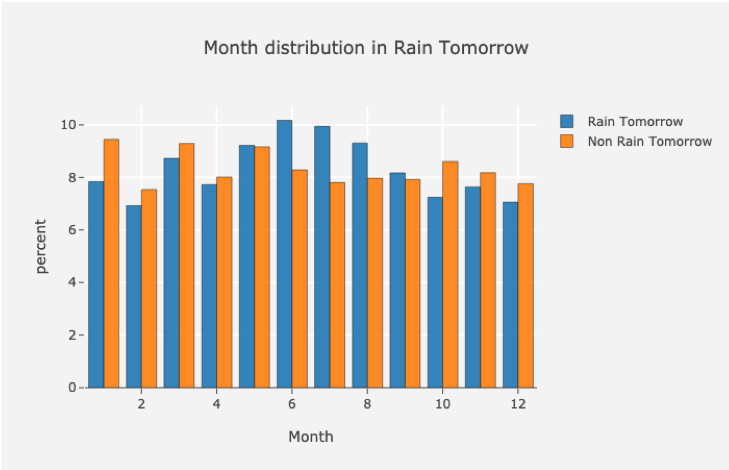}}
\caption{Date (Month) and Rainfall.} 
\end{figure}
\begin{figure}
\centering
\frame{\includegraphics{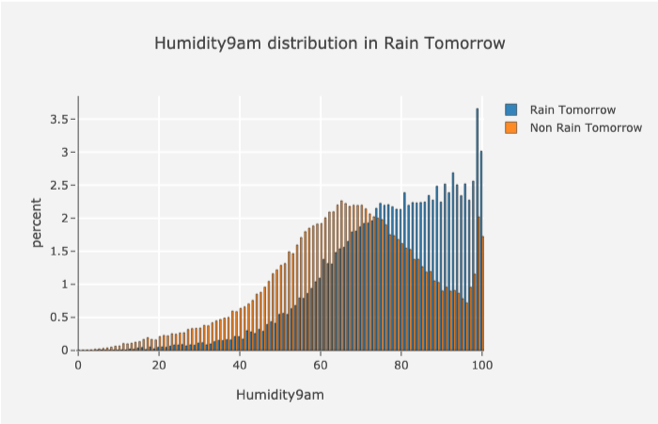}}
\caption{Humidity at 9 AM and Rainfall.} 
\end{figure}

\begin{figure}
\centering
\frame{\includegraphics{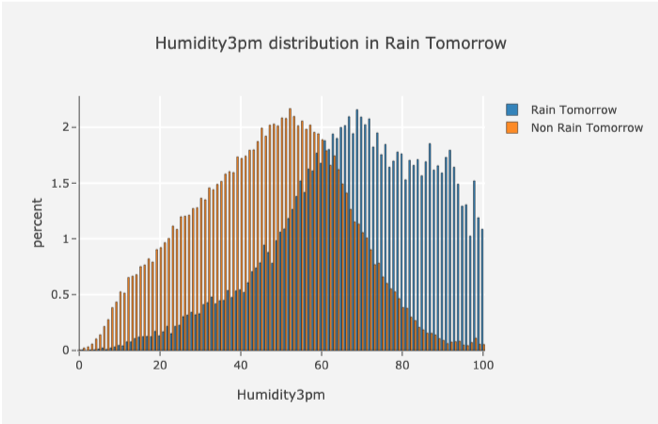}}
\caption{Humidity at 3 PM and Rainfall..} 
\end{figure}
\begin{figure}
\centering
\frame{\includegraphics{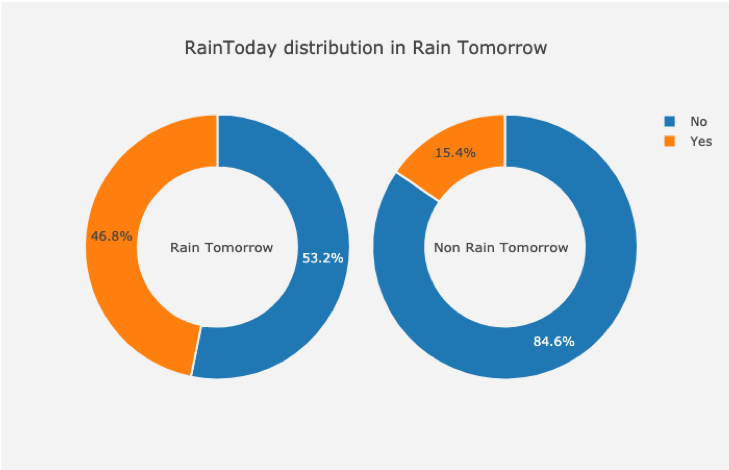}}
\caption{Rain today and Rainfall.} 
\end{figure}

With consideration of domain knowledge, we assumed that rainfall will be strongly related to location and season (date). However, we learned that there is no such pattern with our data. Humidity has a strong correlation with the target variable. However, with further exploration we observed that Humidity at 9 AM and at 3 PM are strongly related and hence we will only consider Humidity at 3 PM. We also observed that Rainfall today and tomorrow are related and hence is a good feature for our predictions. Refer figure 15, 16, 17, 18 and 19 depicting this patterns. \\

\noindent Below are the evaluation metrics for all the experiments carried.

\begin{figure}
\centering
\frame{\includegraphics[width=\columnwidth]{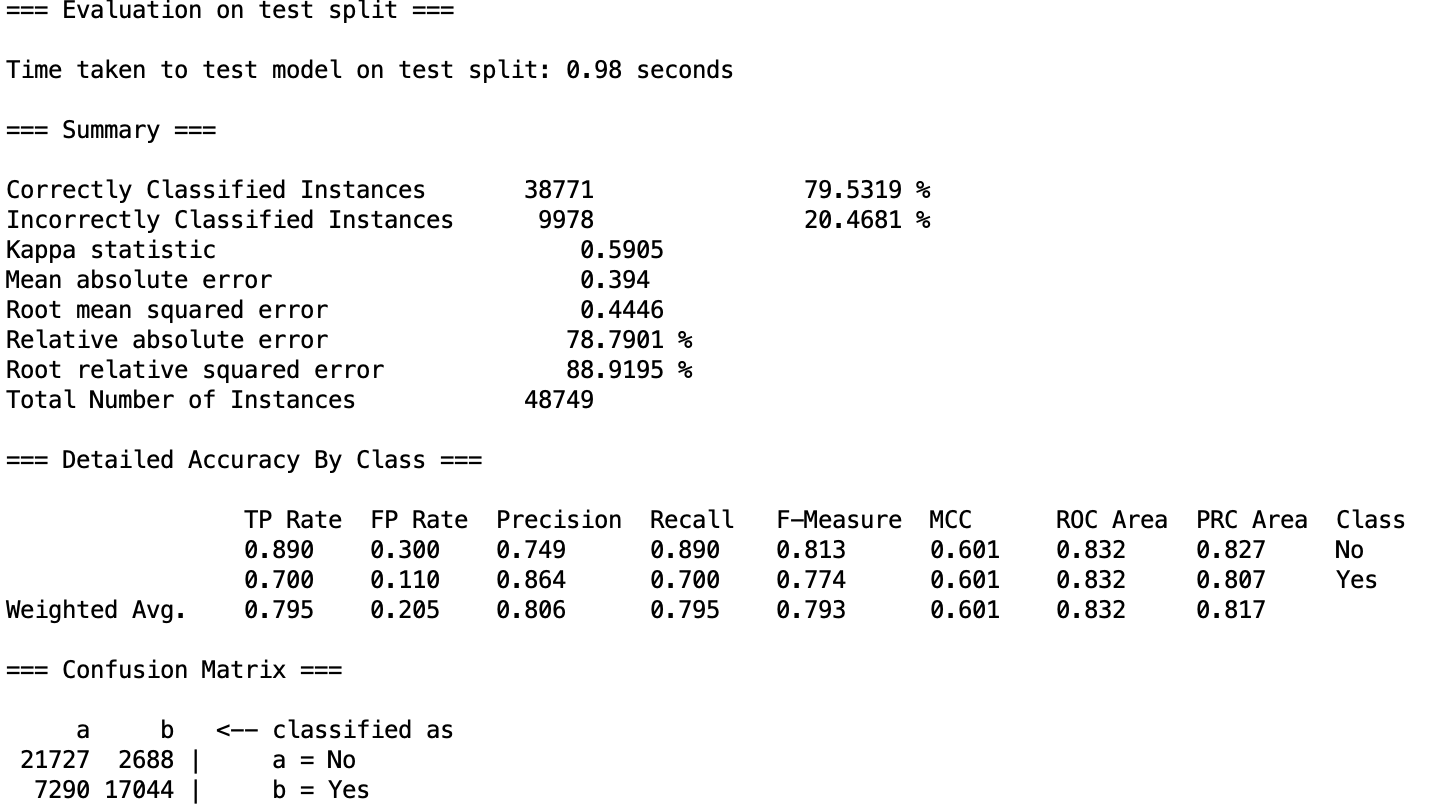}}
\caption{Decision Table using Weka.} \end{figure}
\begin{figure}
\centering
\frame{\includegraphics[width=\columnwidth]{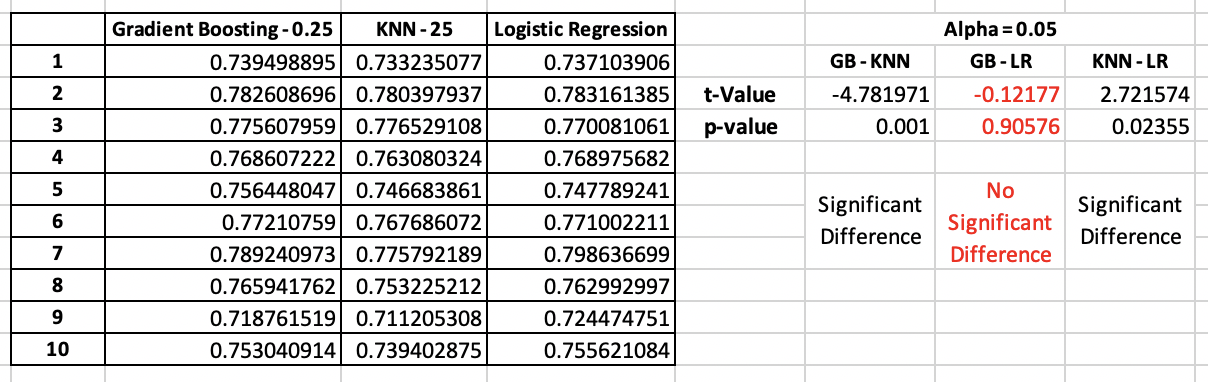}}
\caption{Paired t-test.} 
\end{figure}

\begin{figure}
\centering
\frame{\includegraphics[height=\textheight]{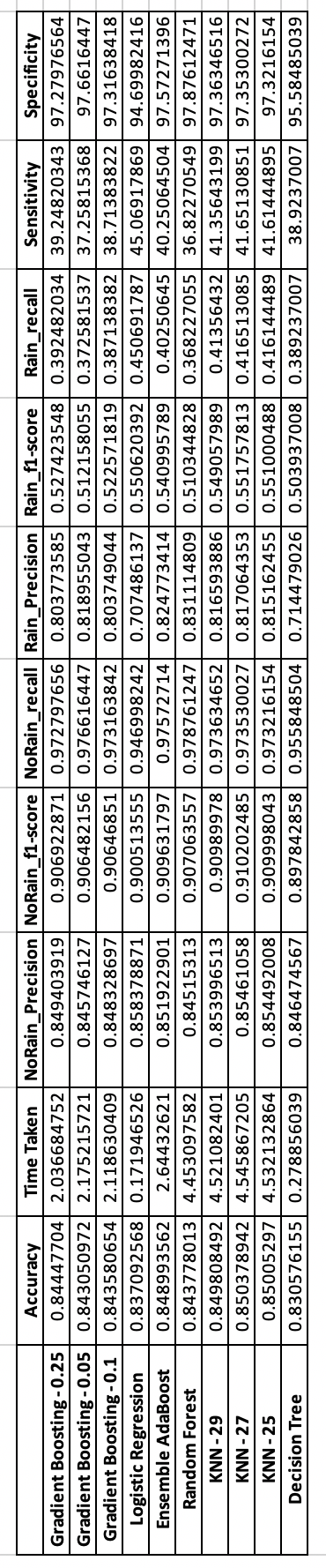}}
\caption{Experiment 1 Evaluation Metrics.} 
\end{figure}

\begin{figure}
\centering
\frame{\includegraphics[height=\textheight]{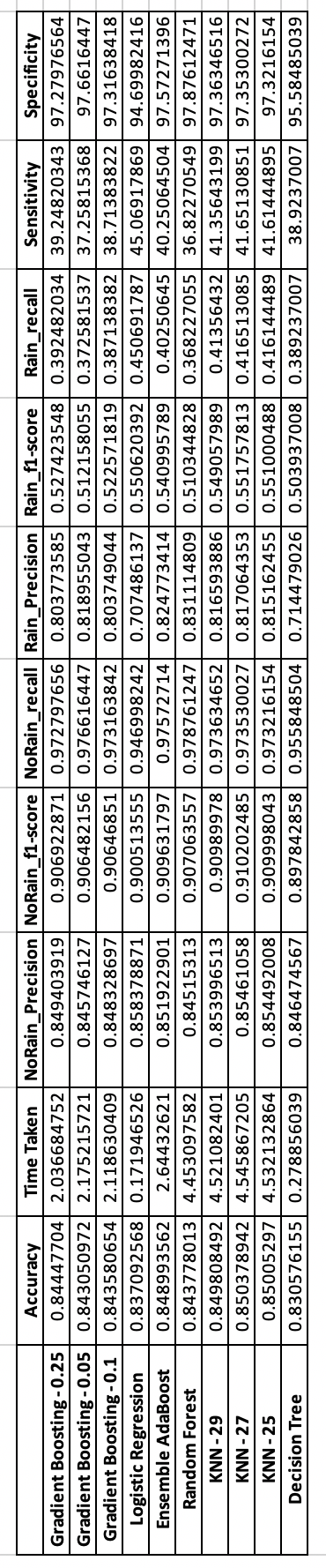}}
\caption{Experiment 2 Evaluation Metrics.} 
\end{figure}
\begin{figure}
\centering
\frame{\includegraphics[height=\textheight]{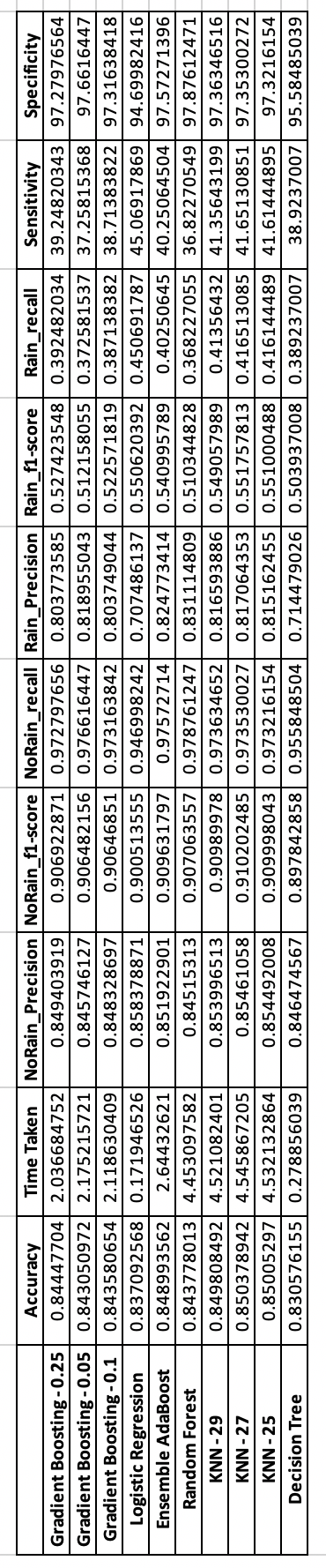}}
\caption{Experiment 3 Evaluation Metrics.} 
\end{figure}

\end{document}